\documentclass[letterpaper, 10 pt, conference]{ieeeconf}  

\IEEEoverridecommandlockouts                              

\usepackage{graphicx}
\usepackage{mathptmx}
\usepackage{times}
\usepackage{amsmath}
\usepackage{amssymb}
\usepackage{color}

\newcommand{\Prob}{\mathbb{P}}

\title{\LARGE \bf
  Time-of-Flight LiDAR-based Precise Mapping
}

\author{Han Wu$^1$ and Zhi Yan$^2$%
  \thanks{$^{1}$School of Mathematical Sciences, Queen Mary University of London, UK.
    {\tt\small han.wu@qmul.ac.uk}}%
  \thanks{$^{1}$CIAD UMR7533, Univ. Bourgogne Franche-Comté, UTBM, F-90010 Belfort, France.
    {\tt\small \{zhi.yan\}@utbm.fr}}%
}

\begin{document}

\maketitle
\thispagestyle{empty}
\pagestyle{empty}

\begin{abstract}
  Last two decades, the problem of robotic mapping has made a lot of progress in the research community.
  However, since the data provided by the sensor still contains noise, how to obtain an accurate map is still an open problem.
  In this note, we analyze the problem from the perspective of mathematical analysis and propose a probabilistic map update method based on multiple explorations.
  The proposed method can help us estimate the number of rounds of robot exploration, which is meaningful for the hardware and time costs of the task.
\end{abstract}

\section{Introduction}
\label{sec:introduction}

Robot mapping, as a fundamental problem in robotics, has been widely studied for many years and considerable progress has been made over the past two decades~\cite{frese05tro,grisetti07tro,endres14tro,krajnik17tro,ls18ral}.
This task requires a robot to explore an environment while building a model of the environment by gathering unknown information, typically acquired by sensors~\cite{yz15iros}.
The model is conventionally represented by a floor plan (e.g. occupancy grid map~\cite{grisetti07tro}) and mainly used for robot navigation, while nowadays contains more dimensional and semantic information~\cite{krajnik17tro,ls18ral} - along with the development of sensors - which can be served for more complex tasks such as 3D navigation, robot manipulation, and long-term mapping.

Nowadays, some applications that are closely related to human life are full of expectations for precise mapping, from the construction and maintenance of general industrial (including exposed pipelines) and living houses, to the search and rescue in mining accidents, urban disasters, hostage situations and explosions.
Among the many off-the-shelf sensors, the Time-of-Flight (ToF) LiDAR should be the most accurate in terms of physical principles.
It measures the distance of objects by sending a laser pulse in a narrow beam towards the object and measuring the time taken by the pulse to be reflected off the target and returned to the sender.
In addition, this type of sensor is robust to lightness variance, making it capable of working in dim environments.
However, the data provided by LiDAR is a sparse set of points, and compared with the image generated by the camera, it lacks easy-to-interpret textures.
Moreover, the laser beam is affected by water droplets and dust in the atmosphere, compared to radar is much more robust.
Of course, these problems can be alleviated by establishing a multi-modal sensor system and multi-sensor data fusion, but these are beyond the scope of this note.

In this note, we want to explore such a problem, that is, given a LiDAR mounted on a mobile robot, how many times the robot needs to perform mapping in a working environment to obtain an environment map that meets a quantified precise demand.
Or, to expand to multi-robot scenarios~\cite{simmons00aaai}, how many homogeneous robots will be needed for collaborative mapping to obtain an environment map that meets the precise demand in a short time.
The expansion of the latter is meaningful, because in search and rescue situations, time is life.

\section{Environmental Exploration}
\label{sec:environmental_exploration}

\subsection{Basic Settings of the Problem}

We (humans, denoted as \textbf{H}) wants a robot \textbf{R} (mounted with a ToF LiDAR) to explore an area \textbf{A} with obstacles, for which there are two possible scenarios: $(1)$ either \textbf{H} has the information on the size and shape of \textbf{A}; $(2)$ or \textbf{H} knows nothing about the size or shape of \textbf{A}.
Case $(1)$ may come when, for example, \textbf{A} is an industrial plant and \textbf{H} wants to know the distribution of equipment and pipelines in \textbf{A} with respect to a plane map of the plant.
Case $(2)$ may come when, for example, \textbf{A} is an old underground mine of which the map is lost.
Obviously, Case $(2)$ is more complicated than Case $(1)$ in the sense that \textbf{R} needs to determine both the size and shape of \textbf{A} and the location of the obstacles.

\textbf{H} may also demand that the map produced by \textbf{R} should be of some precision (e.g. $d = 99\%$).
According to the internal statistical data, \textbf{R} confirms an obstacle with probability $p$ and confirms a free-space surrounding an obstacle with probability $q$.
If a free-space is not surrounding an obstacle, \textbf{R} confirms it surely.
The two parameters $p$ and $q$ stand for the competence of \textbf{R}.
Then the relation between $d$ and $p,q$ also determines different types of the situation: $(A)$ either $d$ is larger than $p$ and $q$, which means the capabilities of \textbf{R} cannot meet the needs of \textbf{H}; or $(B)$ $d$ is smaller than $p$ and $q$.
Case $(B)$ is the normal case that we expect.
Case $(A)$ may come either when the technology is not advanced enough or when the condition of exploration becomes more severe than what a robot can withstand.
This is the case when \textbf{R} should essentially do repeated explorations and use the theory of probability to update (or merge in case of multiple robots) the maps.

Note that in any case, the first exploration of \textbf{R} is to gain knowledge of the shape and size of the environment, denoted as \textbf{M}, i.e. the map of \textbf{A} in which each cell is either an obstacle or a neighbour of an obstacle.
Replacing \textbf{A} by \textbf{M}, Case $(2)$ is reduced to Case $(1)$.
If we are in Case $(B)$, then a map that meets the requirements can be obtained by repeatedly exploring by \textbf{R} and by updating each cell in \textbf{M} using the method of maximum likelihood.

\subsection{Probabilistic Model and Main Questions}

We confine ourselves to Case $(1)+(A)$.
The map \textbf{M} of the environment with obstacles to be explored in our consideration is in a simplified version according to the occupancy grid map.
Namely, it is a matrix $m$ with entries from $\{0,1\}$.
The size and shape of $m$ is assumed to be known to us.
The coordinates of $m$ will be noted by $(x,y)$.
If at $(x,y)$ there is an obstacle, we set $m(x,y)=1$.
If it is free-space, we set $m(x,y)=0$.
We emphasize that $m(x,y)$'s are constants not random variables.

Around an obstacle, \textbf{R} gives a distribution law of detecting errors (denoted as \textbf{DE}) in a neighbourhood (denoted by \textbf{NH}) centered at the obstacle.
The size and shape of \textbf{NH} depends on the distribution law we choose.
It could be a circle (e.g., for a Gaussian distribution) or a square centered at the obstacle, or even anything else.
Note that \textbf{NH} is used at the preliminary step to determine \textbf{M}.
For any $(x,y)$, let $\textbf{NH}(x,y)$ be the neighbourhood centered at $(x,y)$.
Then for any $(x,y) \in \textbf{M}$, there is some $(x',y') \in \textbf{M}$ with $m(x',y')=1$ and $(x,y) \in \textbf{NH}(x',y')$.

Each single exploration, indexed by a positive integer $i \in [1,N]$ has an empty map $m_i$ of the same size as $m$ with initial assignment $m_i(x,y) = -1$ (representing unknown space) at every $(x,y)$.
After the exploration, \textbf{R} gives a matrix $m_i$ with entries from $\{0,1\}$.
Given $(x,y)$, $m_i(x,y)$ for integers $i \in [1,N]$ are i.i.d. random variables, whose distribution law depends on the value of $m(x,y)$.
More precisely, if $m(x,y)=1$, $m_i(x,y)$ obeys a law of Bernoulli distribution with
$$ \Prob(m_i(x,y)=1) = p$$
for some $p$ with $0<p<1$ and close to $1$; if $m(x,y)=0$, $m_i(x,y)$ obeys a law of Bernoulli distribution with
$$ \Prob(m_i(x,y)=0) = q$$
for some $q$ with $0<q<1$ and close to $1$.
Note that, contrary to $p$, which is a constant, $q=q(x,y)$ depends the position and the distribution of possibly all $(x',y') \in \textbf{M}$ with $m(x',y')=1$.
However, given \textbf{DE}, it is possible bound $q$ from below and from above by some constants depending on $p$ using some geometric analysis on all patterns of distributions of obstacles.
We take
\begin{equation}
	q' = \min(q)
\label{ParamRel}
\end{equation}
where $\min(\cdot)$ is taken over all patterns of distributions of obstacles.
The parameter $p$ should be determined by experiments before the exploration and $q'$ is calculated thereafter.
Note that if absolutely nothing is known about the distribution of obstacles (e.g. they should have forms of a line, a circle, etc.) we should take $q' = p$, while in the case we know they form lines sufficiently far away from each other, and if \textbf{NH} has the form of a square of size $3 \times 3$, we can take $q'=1-\frac{3(1-p)}{8}$.

After \textbf{R} has finished $N$ rounds of explorations, we shall fuse the $N$ matrices $m_i$ into one $m_0$ by a method which will be given in Section~\ref{sec:probabilistic_map_update}.
We expect $m_0 = m$ with a given confidence level $d$, i.e. $\Prob(m_i(x,y)=m(x,y)) \geq d$ for any $(x,y)$.
The closer $d$ gets to $1$, the more rounds we need to attain the confidence level $d$.
We focus on the case that $p,q$ are relatively smaller than $d$.
Consequently, we need many rounds to explore every point $(x,y)$ in order to achieve the wanted confidence level $d$.
Based on the above analysis, it is natural to have the following two questions:
\begin{itemize}
    \item For any given $d$, what is the minimal number of rounds $N(d)$ so that our method gives $m_0=m$ with the confidence level $d$?
    \item If the available quantity of rounds $N < N(d)$ (e.g. due to time or battery life limitations), what can we produce out of the maps $m_i$?
\end{itemize}
We will give the answers in the next section.

\section{Probabilistic Map Update}
\label{sec:probabilistic_map_update}

\subsection{Description of the Method}

Our intuition is based on Large Number Theorem.
In our situation, it says that, if $m(x,y)=1$, then the mean value
$$ M(x,y) = \frac{1}{N} \sum_{i=1}^N m_i(x,y) \to p, \text{ as } N \to \infty; $$
and if $m(x,y)=0$, then
$$ M(x,y) = \frac{1}{N} \sum_{i=1}^N m_i(x,y) \to 1-q, \text{ as } N \to \infty. $$
Since $p$ is close to $1$ and $1-q$ is close to $0$, the mean value $M(x,y)$ should be a good quantity to distinguish the two cases $m(x,y)=1$ and $m(x,y)=0$.
More precisely, we expect the existence of some number $C \in (1-q,p)$ and set
\begin{equation}
	m_0(x,y) = \left\{ \begin{matrix} 1 & \text{if } M(x,y) \geq C \\ 0 & \text{if } M(x,y) < C, \end{matrix} \right.
\label{Method}
\end{equation}
such that we have both
\begin{equation}
	\Prob(m_0(x,y)=1) \geq d \text{ if } m(x,y)=1
\label{ObsCond}
\end{equation}
and
\begin{equation}
	\Prob(m_0(x,y)=0) \geq d \text{ if } m(x,y)=0.
\label{NonObsCond}
\end{equation}

\subsection{Determination of the Parameters}

When $N$ is large, it is usually not easy to execute calculation with the exact distribution law of $M(x,y)$.
We may simplify the calculation using Central Limit Theorem, which says in our situation that in the sense of convergence in law
$$ \left\{ \begin{matrix} \frac{M(x,y) - p}{\sqrt{p}} \cdot \sqrt{N} \to \mathcal{N}(0,1) & \text{if } m(x,y) = 1 \\ \frac{M(x,y) - (1-q)}{\sqrt{1-q}} \cdot \sqrt{N} \to \mathcal{N}(0,1) & \text{if } m(x,y) = 0 \end{matrix} \right. $$
where $\mathcal{N}(0,1)$ is the standard Gaussian distribution. We thus simply assume
\begin{equation}
	\left\{ \begin{matrix} \frac{M(x,y) - p}{\sqrt{p}} \cdot \sqrt{N} = \mathcal{N}(0,1) & \text{if } m(x,y) = 1 \\ \frac{M(x,y) - (1-q)}{\sqrt{1-q}} \cdot \sqrt{N} = \mathcal{N}(0,1) & \text{if } m(x,y) = 0. \end{matrix} \right.
\end{equation}
Then (\ref{ObsCond}) is translated into
$$ \int_{\frac{C - p}{\sqrt{p}} \cdot \sqrt{N}}^{\infty} \frac{1}{\sqrt{2\pi}} e^{-\frac{x^2}{2}}dx \geq d, $$
which is equivalent to
\begin{equation}
	\frac{C - p}{\sqrt{p}} \cdot \sqrt{N} \leq a
\label{FObsCond}
\end{equation}
for some $a=a(d)<0$ depending only on $d$.
Similarly, (\ref{NonObsCond}) is translated into
$$ \int_{-\infty}^{\frac{C - (1-q)}{\sqrt{1-q}} \cdot \sqrt{N}} \frac{1}{\sqrt{2\pi}} e^{-\frac{x^2}{2}}dx \geq d, $$
which is equivalent to
\begin{equation}
	\frac{C - (1-q)}{\sqrt{1-q}} \cdot \sqrt{N} \geq b
\label{FNonObsCond}
\end{equation}
for some $b=b(d)>0$ depending only on $d$ ($b=-a$ for example). In order that both (\ref{FObsCond}) and (\ref{FNonObsCond}) can be satisfied, we must have
$$ \frac{b\sqrt{1-q}}{\sqrt{N}}+1-q \leq C \leq \frac{a\sqrt{p}}{\sqrt{N}}+p, $$
hence
$$ \sqrt{N} \geq \frac{b\sqrt{1-q} - a\sqrt{p}}{p-(1-q)}. $$
Note that the right hand side is an increasing function of $1-q$ and by (\ref{ParamRel}) $1-q \leq 1-q'$, we shall thus put $1-q=1-q'$ and take
\begin{equation}
	N(d) = \left\lceil \left( \frac{b\sqrt{1-q'} - a\sqrt{p}}{p+q'-1} \right)^2 \right\rceil.
\label{RNumberLowerBound}
\end{equation}
We take $C$ to be any number satisfying
\begin{equation}
	\frac{b\sqrt{1-q'}}{\sqrt{N}}+1-q' \leq C \leq \frac{a\sqrt{p}}{\sqrt{N}}+p,
\label{CriticalPoint}
\end{equation}

\subsection{Methods when $N < N(d)$}

Lack of time or due to energy budget, it may happen that $N < N(d)$.
In this case, it is no longer possible to completely achieve the asked precision of the produced map.
As a solution, \textbf{R} may employ the method of maximum likelihood.
Note that the number of rounds which assign $1$ to the position $(x,y)$ is $NM(x,y)$.
Then the number of rounds which assign $0$ to the position $(x,y)$ is $N-NM(x,y)$.
If $m(x,y)=1$, then the probability that this obtained assignments happen is
$$ p^{NM(x,y)}(1-p)^{N-NM(x,y)}. $$
If $m(x,y)=0$, then the probability that this obtained assignments happen is
$$ q^{N-NM(x,y)}(1-q)^{N-NM(x,y)}. $$
If $ p^{NM(x,y)}(1-p)^{N-NM(x,y)} \geq q^{N-NM(x,y)}(1-q)^{N-NM(x,y)} $ then we put $m_0(x,y)=1$; otherwise we put $m_0(x,y)=0$.
The problem is, it is not obvious how to calculate the precision of the output map.

Alternatively, we can apply the method described in the first part with some $d'<d$. The best (largest) $d'$ we can get satisfies
$$ d'= \int_{\frac{C - p}{\sqrt{p}} \cdot \sqrt{N}}^{\infty} \frac{1}{\sqrt{2\pi}} e^{-\frac{x^2}{2}}dx = \min \int_{-\infty}^{\frac{C - (1-q)}{\sqrt{1-q}} \cdot \sqrt{N}} \frac{1}{\sqrt{2\pi}} e^{-\frac{x^2}{2}}dx. $$
where $\min $ is taken under (\ref{ParamRel}). By the symmetry of the integrand, this gives
$$ \frac{C - p}{\sqrt{p}} \cdot \sqrt{N} = - \frac{C - (1-q')}{\sqrt{1-q'}} \cdot \sqrt{N} $$
hence
$$ C = \frac{p\sqrt{1-q'}+(1-q')\sqrt{p}}{\sqrt{p}+\sqrt{1-q'}}. $$
We thus get
$$ d' = \int_{\frac{1-p-q'}{\sqrt{p}+\sqrt{1-q'}} \cdot \sqrt{N}}^{\infty} \frac{1}{\sqrt{2\pi}} e^{-\frac{x^2}{2}}dx. $$

\subsection{Description of the Algorithm}

\begin{itemize}
\item[1] We choose a model of the detecting errors for a given ToF LiDAR (mounted on \textbf{R}).
This includes the choice of \textbf{NH} and a probability measure \textbf{DE} on it with mass $p$ at the center.
Calculate $q'$.
Note that if there is absolutely no information about the distribution of obstacles, we take $q'=p$.
\item[2] Set a confidence level $d$ which is relatively larger than $p$.
Find $a$ such that
  $$ \int_a^{\infty} \frac{1}{\sqrt{2\pi}} e^{-\frac{x^2}{2}}dx = d $$
  in a table of statistic book. Set $b=-a$.
\item[3] Calculate $N(d)$ by (\ref{RNumberLowerBound}).
Choose $N \geq N(d)$. Choose $C$ by (\ref{CriticalPoint}).
\item[4] Run explorations until we get $N$ maps $m_i$.
Use (\ref{Method}) to fuse the maps into one $m_0$.
\end{itemize}

\section{Conclusions}
\label{sec:conclusions}

In this note, we analyzed how to obtain an accurate robot-made map from the perspective of mathematical analysis, and introduced a probabilistic map update method.
This method is designed based on the repeated mapping of the same environment by a robot, and can help us estimate the number of exploration rounds, while the latter is meaningful for the cost estimation of mapping tasks.

\bibliographystyle{IEEEtran}
\bibliography{main}

\end{document}